\documentclass[conference]{IEEEtran}
\IEEEoverridecommandlockouts
\usepackage{cite}
\usepackage{amsmath,amssymb,amsfonts,bm}
\usepackage{algorithmic}
\usepackage[ruled,norelsize]{algorithm2e}
\usepackage{setspace}
\usepackage{graphicx}
\usepackage{textcomp}
\usepackage{xcolor}
\usepackage{caption}
\usepackage{subcaption}
\usepackage[detect-weight]{siunitx}
\newcommand{\vect}[1]{\boldsymbol{#1}}
\makeatletter
\newcommand{\removelatexerror}{\let\@latex@error\@gobble}
\makeatother

\def\IEEEtitletopspace{18pt}
\SetKw{Continue}{continue}

\def\BibTeX{{\rm B\kern-.05em{\sc i\kern-.025em b}\kern-.08em
    T\kern-.1667em\lower.7ex\hbox{E}\kern-.125emX}}
\begin{document}

\title{Deadlock-Free Collision Avoidance for Nonholonomic Robots}

\author{\IEEEauthorblockN{1\textsuperscript{st} Ruochen Zheng}
\IEEEauthorblockA{\textit{Dept. of Modeling, Optimization and Simulation} \\
\textit{Megvii Automation \& Robotics}\\
Beijing, China \\
zhengruochen02@megvii.com}
\and
\IEEEauthorblockN{2\textsuperscript{nd} Siyu Li}
\IEEEauthorblockA{\textit{Dept. of Modeling, Optimization and Simulation} \\
\textit{Megvii Automation \& Robotics}\\
Beijing, China \\
lisiyu02@megvii.com}
}

\maketitle

\begin{abstract}
We present a method for deadlock-free and collision-free navigation in a multi-robot system with nonholonomic robots. The problem is solved by quadratic programming and is applicable to most wheeled mobile robots with linear kinematic constraints. We introduce masked velocity and Masked Cooperative Collision Avoidance (MCCA) algorithm to encourage a fully decentralized deadlock avoidance behavior. To verify the method, we provide a detailed implementation and introduce heading oscillation avoidance for differential-drive robots. To the best of our knowledge, it is the first method to give very promising and stable results for deadlock avoidance even in situations with a large number of robots and narrow passages. 
\end{abstract}


\section{Introduction} \label{intro}

Collision and deadlock avoidance in a multi-robot system is a fundamental problem to solve in order to put such systems into application. While a centralized system can simultaneously determine the navigation details for all the robots, curse-of-dimensionality will prevent the system from giving a satisfying solution in acceptable time interval for even only a moderate number of robots. As a result, we believe that it is more promising to search for a decentralized navigation strategy where each robot will fully utilize its accessible information to make decisions for its own benefits \cite{sheffi1984}.

Many works with decentralized reciprocal navigation strategy use a velocity obstacle to represent a time-space area for the velocity to avoid. First conceptualized in \cite{fiorini1998motion}, velocity obstacles make it easy to represent velocity constraints mathematically. A promising framework, Optimal Reciprocal Collision Avoidance (ORCA) \cite{berg2011reciprocal}, is shown to be able to solve collision avoidance geometrically and optimally. For each robot, a linear programming (LP) problem minimizes the diversion of the chosen velocity to the preferred velocity, with permitted velocities constructed as ORCA half-planes. The maximum distance to violated half-planes is minimized if there is no valid solution. While reciprocal decision-making mechanism is able to guarantee collision avoidance, it is usually not able to avoid or resolve deadlock. The situation is worse with dense obstacles presented in the environment. 

Inspired by velocity obstacle and ORCA framework, we propose a quadratic programming (QP) formulation to solve collision avoidance problems for nonholonomic robots with linear kinematic constraints. The formulation is a generalized form with its holonomic case similar to the original ORCA problem. The formulation can be applied to most controllable wheeled mobile robots in the real world with linear kinematic models. 

Moreover, this paper introduces masked velocity and Masked Cooperative Collision Avoidance (MCCA) half-plane to encourage deadlock-free navigation. A temporary and local cooperative behavior is motivated by escalating the priority of certain robots and use masked velocity as the deadlock-free velocity intention a robot updates and tries to approach. The propagation of masked velocity among robots will gradually influence the real velocity of other robots and encourage deadlock-free navigation continuously. 

We provide an application of the QP formulation and MCCA method using differential-drive robots. It is shown to be able to solve the collision avoidance problem efficiently and the experimentation confirms that masked velocity, MCCA half-planes and robot priority determination combined are able to resolve deadlock in various scenarios reliably with a low level of local cooperation among robots. It also applies to robots with low acceleration capability and difficult to maneuver. Besides, the formulation provides robots with better mobility in narrow passages as there is no need to enlarge the radius of robots significantly. 

The rest of this paper is organized as follows. A brief summary of related work is provided in Section \ref{review}. We present the general QP formulation for solving nonholonomic collision avoidance problems in Section \ref{qp}. In Section \ref{mcca}, we introduce masked velocity and MCCA half-planes and give the algorithms for robot prioritization and deadlock-free navigation. We combine the QP formulation and MCCA to give a working example in Section \ref{example} for differential-drive robots and report simulation results in Section \ref{exp}.

\section{Related Work}\label{review}
We briefly review related work in velocity obstacles and optimal reciprocal collision avoidance. 

As an early effort, \cite{fiorini1998motion} conceptualizes Velocity Obstacle (VO), which refers to circular robots and static obstacles. VO makes it easy to represent the velocity constraints mathematically. Later in\cite{Berg2008rvo}, Reciprocal Velocity Obstacle (RVO) is proposed, which is an extension of the original VO and solves the oscillation problem induced by VO. Basically, RVO keeps the VO stable from velocity change, such that the original velocity will not be chosen later. It also proves that RVO is complete from inducing collisions. Another effort to improve VO is \cite{guy2009clearpath}, which defines FVO, allowing a robot to avoid others but only in a pre-determined time interval. The resulted VO is a truncated cone. FVO makes solution space much larger compared to the original VO as it is only necessary to avoid collision for a short time interval. 

ORCA \cite{berg2011reciprocal} utilizes the advantage of both RVO and FVO to solve a linear programming (LP) problem for each robot in the system to determine its optimized velocity. Robots are assumed to be holonomic. For each robot, a LP problem to minimize the diversion of the chosen velocity to the preferred velocity (given from a global planner), with permitted velocities constructed with ORCA half-planes.\footnote{As suggested in the original ORCA, it is ideal to set $\vect{v}^{\text{opt}} = \vect{v}^{\text{current}}$, which applies to this paper unless specified otherwise.} Given the unique structure of the problem, the LP can be easily solved with an incremental insertion of constraints and is guaranteed to be global optimal. Two most referred challenges of the original ORCA are mentioned below.

The first challenge is brought by the holonomic assumption. Most robots in real-world applications are nonholonomic. The solved velocity may not be reached directly and instantly, leaving a gap between the target trace and the realized trace, which could result in collision. A walk-around solution to incorporating kinematics of differential-drive robots is provided in \cite{snape2010smooth}, which adopts the kinematic adaption approach described in \cite{Kluge2005} and introduces effective center with fully controlled velocity at the cost of a larger disc radius. However, it uses a significantly larger effective radius and does not explore the cases where the translational distance $D$ is much smaller. A similar approach (NH-ORCA) is presented in \cite{Alonso-Mora2013}, where the radius of a nonholonomic robot is enlarged by a tracking error $\epsilon$ between the perfect holonomic motions and nonholonomic motions. Some recent efforts for nonholonomic collision avoidance considering RVO can be found in \cite{alonso2018} and \cite{sainte2020}.

The second challenge is about the occurrence of deadlock and oscillation in a multi-robot system, where two or more robots are stuck together and unable to resolve the situation for a long time period. While a centralized decision-making mechanism (see for example \cite{cap2016}, \cite{der2020} and \cite{der2021}) may eventually solve the problem, it is too time-consuming and unstable to be applied to situations with only a moderate number of robots; Moreover, the evaluation and simulation of such mechanism in a discrete time-space is questionable. Some decentralized efforts can be found as well, such as \cite{arul2021} and \cite{wang2020}. However these mechanisms are mostly not able to deal with narrow passages. Ideally, the level of global coordination among robots should be minimum yet effective to avoid deadlock and oscillation with various scenarios in real time, and we shall pursue a deadlock-free method in such direction.


%

\section{Quadratic programming formulation}\label{qp}
Compared to the LP formulation in ORCA, we would like a collision avoidance formulation to have the following key features.

\begin{itemize}
\item it is able to directly optimize the translational velocity $\vect{v} = \left(v_x, v_y \right)$ to be close to $\vect{v}^{\text{pref}}$. 
\item If it is not possible to meet all the half-plane constraints, the violation should be minimized.
\item Control inputs and nonholonomic kinematic constraints can be incorporated directly.
\end{itemize}

A corresponding quadratic programming problem can be formulated by 
\begin{subequations}\label{Gform:main}
\begin{align}
& \text{min}  &&  \alpha _{1} \left \| \vect{v}-\vect{v}^{\text{pref}} \right \| _{2}^{2} + \alpha _{2}\sum_{i\in \mathcal{O}}^{} \delta_{i}^{2}  +\alpha _{3}\sum_{j\in \mathcal{R}}^{} \delta_{j}^{2} &   & \tag{\ref{Gform:main}} \\
& \text{s.t.} && \det\left( \bm{B} \right) \leq \delta_{\zeta}, \zeta\in \mathcal{O}\cup \mathcal{R} ,\delta_{\zeta}\geq 0  \label{Gform:ca}  \\
&                && \dot{\vect{q}}=\bm{S}(\vect{q})\cdot\vect{r} \label{Gform:kin1} \\
&                && \begin{bmatrix}  \bm{F}&\bm{0} \\ \bm{0}&\bm{G} \end{bmatrix}\begin{bmatrix} \dot{\vect{q}} \\ \vect{r}  \end{bmatrix} = 0.   \label{Gform:kin2} 
\end{align}
\end{subequations}

\begin{figure}[tb]
\centerline{\includegraphics[scale=0.15] {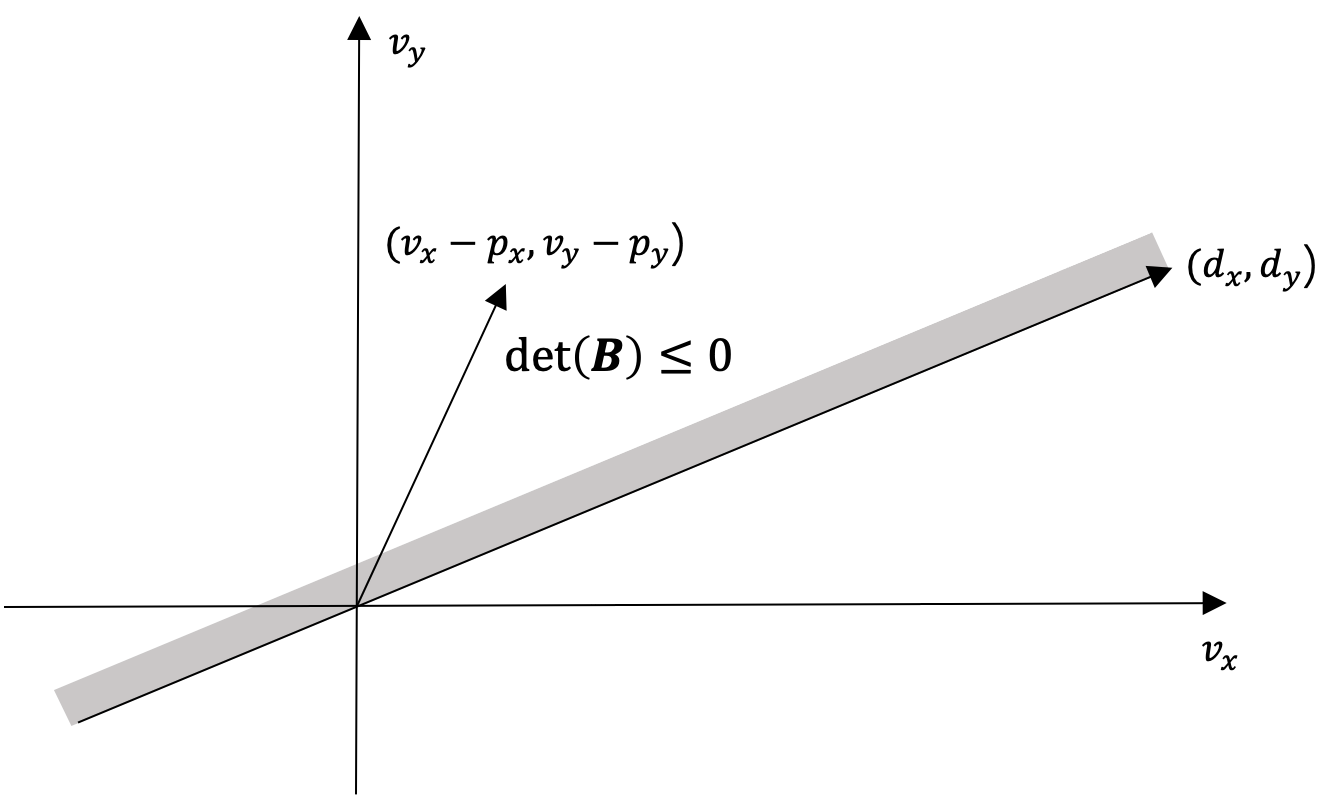}}
\caption{An example of a half-plane and its permitted velocities. A permitted upper half-plane is described by position $\left ( p_{x}^{\zeta}, p_{y}^{\zeta} \right)$ and direction$\left ( d_{x}^{\zeta}, d_{y}^{\zeta} \right)$ in velocity space. To translate position $\left ( p_{x}^{\zeta}, p_{y}^{\zeta} \right)$ into the origin point, the velocity vector becomes $\left( v_{x} - p_{x}^{\zeta},v_{y} - p_{y}^{\zeta} \right)$. The determinant of $\bm{B} =\begin{bmatrix}v_{x} - p_{x}^{\zeta}&d_{x}^{\zeta} \\[1pt]v_{y} - p_{y}^{\zeta} &d_{y}^{\zeta} \end{bmatrix} $ should be nonpositive if the velocity lies inside or on the upper half-plane thus permitted.}
\label{fig-ca-constraint}
\end{figure}

Constraint \eqref{Gform:ca} describes the half-planes of permitted velocities for the current robot. Each of the half-plane is induced by a robot $\zeta\in \mathcal{R}$ or a static obstacle $\zeta\in \mathcal{O}$. Fig. \ref{fig-ca-constraint} explains the relationship between a half-plane and its permitted velocities. Normally, the collision avoidance constraint would request $\det\left( \bm{B} \right) \leq 0$, which could be difficult to achieve in crowded situations. We introduce an auxiliary variable $\delta_{\zeta}\geq 0$ for each of the constraint in \eqref{Gform:ca} and punish positive auxiliary variables heavily in the objective function, such that the collision avoidance constraints are always met with a minimized velocity violation. 

Constraint \eqref{Gform:kin1} and \eqref{Gform:kin2} incorporate nonholonomic kinematic constraints. Nonholonomic constraints limit possible velocities of the robot or directions of motion \cite{klancar2017}. Consider a robot with motion states described in the generalized coordinates $\vect{q}=[q_{1},\cdots ,q_{n} ]^{T}$. $\dot{\vect{q}}=[\dot{q_{1}},\cdots ,\dot{q_{n}} ]^{T}$ is the corresponding velocity vector. We assume that the robot has $m$ independent motion control motors or other mechanisms, written as $\vect{r}=[r_{1},\cdots ,r_{m} ]^{T}$. For example, the left wheel velocity and right wheel velocity are the two independent motion control variables for a differential-drive robot. The deduction of velocity from the control vector (control inputs) is called forward state estimation or direct kinematic model, which holds for each time instance. In our formulation \eqref{Gform:kin1}, it is assumed that the relationship is linear and $\bm{S}(\vect{q}) = \left[\vect{s}_{1}(\vect{q}),\cdots ,\vect{s}_{m}(\vect{q})\right]$ is a matrix of reachable motion vectors. A linear combination of these vectors by the control vector results in the velocity in the generalized coordinates. The linear assumption is true for most controllable wheeled mobile robots with different kinematic models, such as differential-drive, tricycle, Ackermann-drive, and robot with a trailer\cite{klancar2017}. Therefore, the proposed formulation is applicable to most mobile robots in real-world applications. Constraint \eqref{Gform:kin2} represents customized linear constraints for the velocity vector or the control inputs vector where $\bm{F}$ and $\bm{G}$ are the matrices for parameters. The need for customized constraints may come from dynamic and acceleration limitation, navigation strategy, regulation, etc., which increases the versatility of the formulation.

The first part of the objective function is the square of distance between $\vect{v} = \left(v_x, v_y \right)$ and $\vect{v}^{\text{pref}}$. $v_x$ and $v_y$ are part of the velocity vector $\dot{\vect{q}}$. $\vect{v}^{\text{pref}}$ is given by a global planner. The second part is the penalty for half-plane violation. The QP objective should minimize the first part on condition that the violation to half-planes is none or minimum. Thus, we let $\alpha_2, \alpha_3\gg\alpha_1 >0$ in practice. The proposed formulation is a well-defined convex quadratic programming (CQP) problem, which can be solved efficiently with QP solvers in less than one millisecond. In comparison, the general method proposed in \cite{bareiss2015} requires extensive numeric computation and may fail to find a feasible solution.

If kinematics constraints \eqref{Gform:kin1} and \eqref{Gform:kin2} are removed from the formulation, it would solve collision avoidance for holonomic robots; Moreover, if ORCA half-planes are applied to Constraint \eqref{Gform:ca}, we say that the holonomic QP formulation tries to approach $\vect{v}^{\text{pref}}$ with ORCA half-planes\footnote{To generate half-planes in ORCA, while a robot has a reciprocal permitted half-plane with another robot following shared responsibility principle, the half-plane between the robot and a static obstacle is different and the robot should take full responsibility for collision avoidance.}, which is similar to the linear formulation in ORCA.



\section{Masked cooperative collision avoidance }\label{mcca}
Another challenge brought by reciprocal collision avoidance behavior or shared responsibility could result in deadlock or oscillation in crowded situations or with narrow passages. Deadlock avoidance is a global objective that will fail if no robot is to divert from its original path. Therefore, some robots should be compromised for the greater good. Although centralized decision-making is encouraged in such situation, it could easily result in a time-consuming solving process. In this section, we demonstrate how masked velocity, MCCA half-planes and robots priority determination are combined to avoid and resolve possible deadlocks quickly with fully decentralized decision-making and velocity planning process.

\subsection{Masked  Velocity and MCCA}

We define masked velocity as the deadlock-free intention that a robot updates and tries to approach, which is also public to all other robots in the system. The updating of the intention is influenced by the state that a robot is either avoiding or being avoided by other robots. Assume that all the robots $\mathcal{R}$ can be labeled with either high or normal priority, $A\in \overline{\mathcal{R}}$ and $B\in \underline{\mathcal{R}}$ represent a head robot with high priority (to be avoided) and a normal robot with normal priority (to avoid), respectively. A robot $A\in\overline{\mathcal{R}}$ is able to update its masked veloctiy $\vect{v}_{A}^{\overline{\text{m}}}$ without considering the masked velocity of other robots while a robot $B\in \underline{\mathcal{R}}$ has to take the masked velocity of other robots into consideration to solve its masked velocity $\vect{v}_{B}^{\underline{\text{m}}}$. Let 
\begin{equation}
\vect{v}_{A}^{\overline{\text{m}}}=\underset{\vect{v}^{\text{m}}\in ORCA_{A|\mathcal{O}}^{\tau}}{\mathrm{argmin}}\left \| \vect{v}^{\text{m}}-\vect{v}_{A}^{\text{pref}} \right \|, \label{eq:highmask}
\end{equation}where $ORCA_{A|\mathcal{O}}^{\tau}=\underset{o\in \mathcal{O}}{ \bigcap} ORCA_{A|o}^{\tau}$. And let 
\begin{equation}
\vect{v}_{B}^{\underline{\text{m}}}=\underset{\vect{v}^{\text{m}} \in MCCA_{B}^{\tau}}{\mathrm{argmin}}\left \| \vect{v}^{\text{m}}-\vect{v}_{B}^{\text{pref}} \right \| , \label{eq:lowmask}
\end{equation}where $MCCA_{B}^{\tau}= ORCA_{B|\mathcal{O}}^{\tau} \cap \underset{C\in \mathcal{R}}{ \bigcap} MCCA_{B|C}^{\tau}$. Note that masked velocity is not bounded by $D(0,v_{B}^{\text{max}})$ as the original ORCA does. A larger masked velocity would result in more efficient deadlock resolving behavior.   

Head masked velocity $\vect{v}_{A}^{\overline{\text{m}}}$ in Equation \eqref{eq:highmask} is solved by a holonomic QP (as mentioned in Section \ref{qp}) to approach $\vect{v}^{\text{pref}}$ with ORCA half-planes for static obstacles $ORCA_{A|\mathcal{O}}^{\tau}$, which describes the intention that a head robot $A\in\overline{\mathcal{R}}$ tries to approach its current $\vect{v}^{\text{pref}}$ regardless of all other robots.

Normal masked velocity $\vect{v}_{B}^{\underline{\text{m}}}$ in Equation \eqref{eq:lowmask} is solved by a holonomic QP to approach $\vect{v}^{\text{pref}}$ with MCCA half-planes $MCCA_{B}^{\tau}$ considering static obstacles and all other robots, which describes the intention that a normal robot $B\in \underline{\mathcal{R}}$ could only hope to approach its preferred velocity on condition that it avoids all the static obstacles $ORCA_{B|\mathcal{O}}^{\tau}$ and the masked velocities of all other robots.

Masked velocity is solved by holonomic QP for two reasons. First, masked velocity is not the real velocity to derive control inputs thus not constrained by kinematic constraints. Instead, it is used for masked velocity propagation and influence the real velocity of other robots. Second, without considering the kinematic constraints, the masked velocity solved is often larger and usually takes more time steps to achieve, which strengthens the deadlock-free intention of all the other robots in both duration and magnitude that a normal robot has to consider.

The masked velocity of a normal robot $B$ is influenced by other robots in both $\overline{\mathcal{R}}$ and $\underline{\mathcal{R}}$. In Equation \eqref{eq:mcca}, we propose the following Masked Cooperative Collision Avoidance (MCCA) half-plane to represent the permitted masked velocity half-plane for $B$ against another robot $C$:
\begin{equation}
MCCA_{B|C}^{\tau}=\left \{ \vect{v}^{\underline{\text{m}}}|(\vect{v}^{\underline{\text{m}}}-(\vect{v}_{B}^{\text{opt}}+\vect{u}_{\underline{\text{m}}})) \cdot \vect{n}_{\underline{\text{m}}}\geq 0\right \}, C\in \mathcal{R}. \label{eq:mcca}
\end{equation}

\begin{figure}[tb]
\centerline{\includegraphics[width=0.44\textwidth] {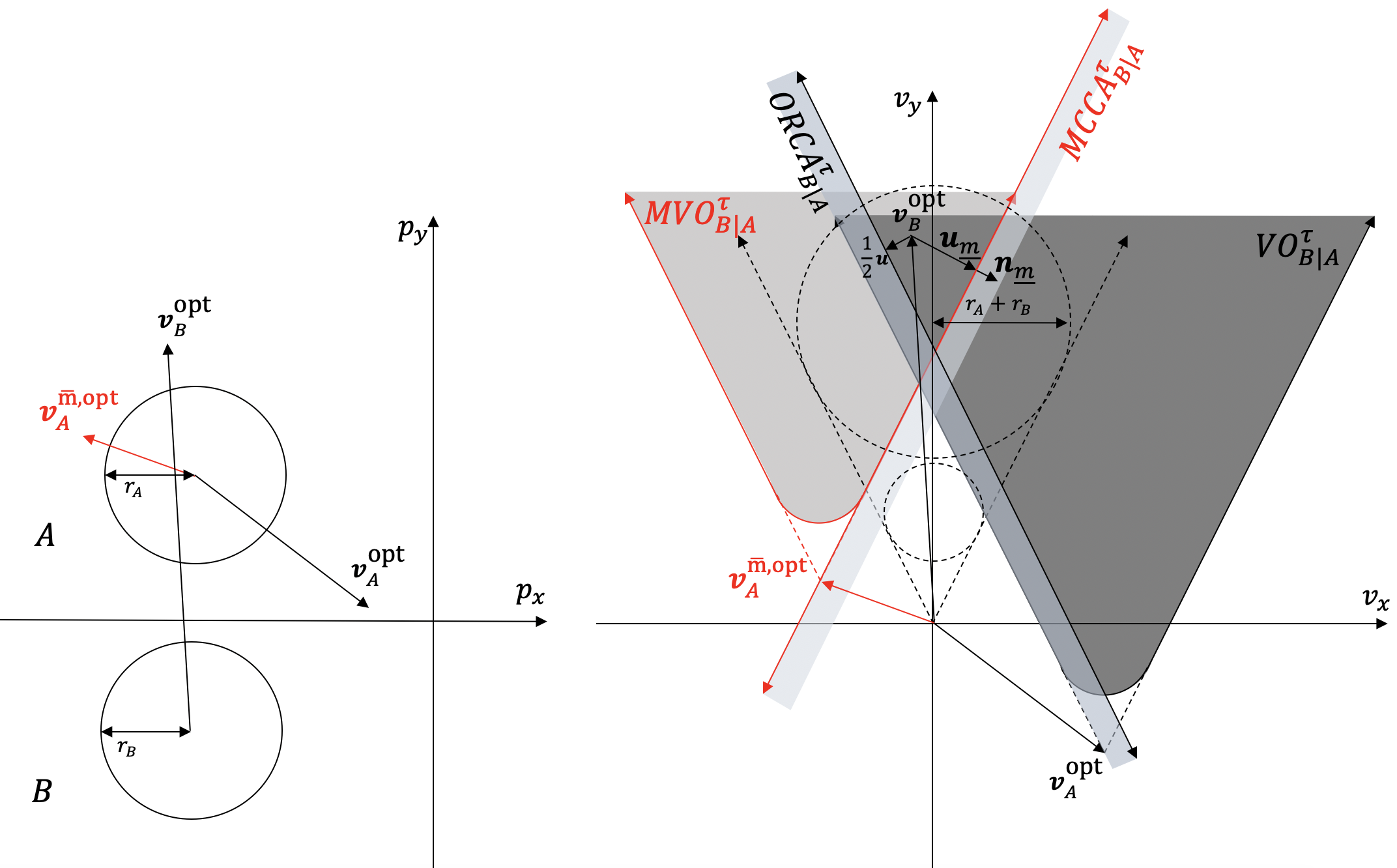}}
\caption{Demonstration of MVO and MCCA half-plane. For $B \in \underline{\mathcal{R}}$, its VO against $A\in \overline{\mathcal{R}}$ is built with $(\vect{v}_A^{\text{opt}},\vect{v}_B^{\text{opt}})$ and shared responsibility principle between $A$ and $B$ is applied (thus the $\frac{1}{2}\vect{u}$) and results in the $ORCA_{B|A}^{\tau}$ half-plane. The MVO for $B$ against $A$, $MVO_{B|A}^{\tau}$, is constructed with $(\vect{v}_A^{\overline{\text{m}}, \text{opt}},\vect{v}_B^{\text{opt}})$. $\vect{u}_{\underline{\text{m}}}$ is the vector from $\vect{v}_{B}^{\text{opt}}-\vect{v}_{A}^{\overline{\text{m}},\text{opt}}$ to the closest point on the boundary of $MVO_{B|A}^{\tau}$, and $\vect{n}_{\underline{\text{m}}}$ is the outward normal of this boundary at point $(\vect{v}_{B}^{\text{opt}} - \vect{v}_{A}^{\overline{\text{m}},\text{opt}}) + \vect{u}_{\underline{\text{m}}}$. $MCCA_{B|A}^{\tau}$ is built with $B$ adapting $\vect{u}_{\underline{\text{m}}}$ alone.}
\label{fig:mcca-a}
\end{figure}

\begin{figure}[tb]
\centerline{\includegraphics[width=0.44\textwidth] {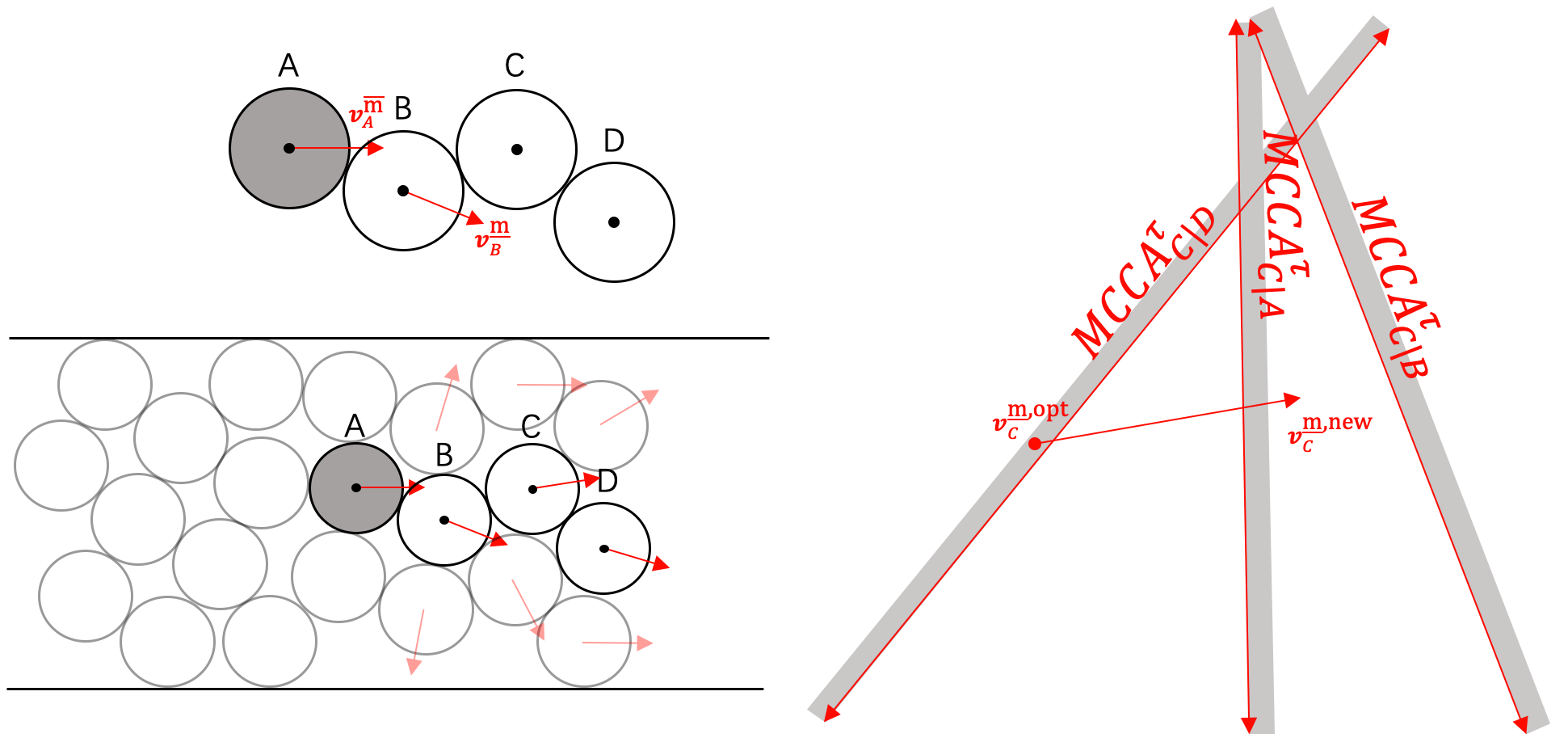}}
\caption{Propagation of masked velocity in a deadlock situation. The masked velocity of head robot $A$ has propagated to normal robot $B$, and normal robot $C$ is being propagated at this time instance. When normal robot $D$ is propagated in the near future, $MCCA^{\tau}_{C|D}$ will move rightward, which will enlarge the masked velocity of $C$. We can describe the process as the deadlock avoidance intention of a head robot propagating forward to normal robots from the near to the distant and the latter making room for the head robot and early propagated normal robots (propagating backward).}
\label{fig:mcca-b}
\end{figure}

Fig. \ref{fig:mcca-a} demonstrates the concept of Masked Velocity Obstacle (MVO) and MCCA half-plane for normal robots. For each control cycle, a robot will update its masked velocity and MCCA half-planes against all other robots. To derive the real collision avoidance velocity of this robot, we adjust the general QP objective function to 

\begin{equation}\label{eq:qp-mcca}
 \text{min} ~\alpha _{1} \left \| \vect{v}-\vect{v}^{\text{pref}} \right \| _{2}^{2} + \alpha _{2}\sum_{i\in \mathcal{O}}^{} \delta_{i}^{2} +\alpha _{3}\sum_{j\in \mathcal{R}}^{} \delta_{j}^{2} + \alpha _{4}\sum_{k\in \mathcal{M}}^{} \delta_{k}^{2},
\end{equation}
where $\mathcal{M}$ is the set of MCCA half-planes against all other robots, $\mathcal{O}$ and $\mathcal{R}$ are the set of ORCA half-planes against obstacles and all other robots, respectively. As mentioned earlier, $\mathcal{M}=\emptyset$ for a head robot since it is not required for the robot to avoid the masked velocity of other robots. Thus the objective in Equation \eqref{Gform:main} is still true for head robots. For each normal robot, it is required that $\alpha_2,\alpha_3\gg\alpha_4\gg\alpha_1$ so that the most prioritized target is to remain clear from static obstacles and other robots, then it is necessary to avoid the masked velocity of all other robots to carry out the intention to avoid deadlock, and finally it could catch up with its preferred velocity. At each control cycle, the masked velocity of all other robots may influence the real velocity of the current robot via MCCA half-planes, and when the control cycle is ended for this robot with updated masked velocity and real velocity, the masked velocity and real velocity of other robots nearby can be influenced in their future control cycles. Consequently, the propagation of masked velocity or deadlock-free intention among robots is possible in both time and space, as shown in Fig. \ref{fig:mcca-b}.

The $\vect{u}_{\underline{\text{m}}}$ is not divided by $2$ in Equation \eqref{eq:mcca} to make sure robot $B$ with normal priority is fully responsible for avoiding the masked velocity of a head robot. Moreover, it encourages the forward and backward propagation of masked velocities among robots with normal priority. Consider the coefficient for $\vect{u}_{\underline{\text{m}}}$ is $0<\beta<1$, each time the masked velocity of head robot $A$ is propagated to a group of normal robots $\mathcal{L}$ not influenced by MVOs of earlier propagated normal robots, the translation of MCCA half-planes of robots in $\mathcal{L}$ will be decreased by a ratio of $\beta$, suggesting that the forward propagation will diminish too soon. Conversely, the backward propagation is affected as well because to avoid the head robot, those early propagated normal robots may need more space that $\mathcal{L}$ fails to provide because masked velocity of robots in $\mathcal{L} $ is hardly influenced.
 
\subsection{Robot Priority Determination}

\setlength{\belowcaptionskip}{-10pt}
\begin{figure}[tb]
 \removelatexerror
  \begin{algorithm}[H]
    \small
    \setstretch{0.90}
   \caption{Robot Priority Determination for $A_{i}$}
    \hspace*{\algorithmicindent} \textbf{Input} 
    $\mathcal{R}$: All robots $A_{i}$: Current robot \\
    \hspace*{\algorithmicindent} \textbf{Output} 
   Updated priority for $A_{i}$\\
    \textit{Initialization} $T_{A_{i} }=0$, $S_{A_{i} }=0$, $A_{i} \in\underline{\mathcal{R}}$\\
	 \eIf{($A_{i}$ reaches its goal)}{
	 	Set $A_i$ as a normal robot, $T_{A_{i} }=0$, $S_{A_{i} }=0$\;
	 }{
	 	\eIf{($T_{A_{i} }>0$)}{
			Set $A_i$ as a normal robot,  $T_{A_{i} }=T_{A_{i} } - 1$\;
		}{
			\eIf{($\exists~ A_{j} \in \overline{\mathcal{R}}$ s.t. $\vect{v}_{A_{i}}^{{\normalfont \overline{\text{m}}}} \in MVO_{A_{i}|A_{j}}^{\infty}$ and $\vect{v}_{A_{i}}^{{\normalfont \overline{\text{m}}}}\cdot \vect{v}_{A_{j}}^{{\normalfont \overline{\text{m}}}}<0$ and $S_{A_{i} }<S_{A_{j}}$)}{
			Set $A_i$ as a normal robot, $T_{A_{i} }=\eta$\;
		}{
			Set $A_i$ as a head robot, $S_{A_{i} }=S_{A_{i} }+1$
		}
	 }
	 }
\end{algorithm}
\caption{Algorithm for robot priority determination. $\eta$ is a preset integer representing the initial tabu steps for a newly determined normal robot conflicting with other head robots. The robot's chance to raise priority is postponed until the remaining tabu steps $T_{A_{i}}$ is zero. $S_{A_{i}}$ is the number of control cycles where $A_i$ is of high priority by far, which is an accumulative indicator to label the importance of $A_i$. Both $T_{A_{i}}$ and $S_{A_{i}}$ are reset to zero once the robot reaches its current goal.}
\label{fig:priority-alg}
\end{figure}

We adopt a fully decentralized method for each robot to determine its priority autonomously at each time step before masked velocity can be applied. Fig. \ref{fig:priority-alg} presents the algorithm for robot priority determination. Since the masked velocity of a head robot only tries to avoid ORCA half-planes of obstacles, the conflict of masked velocity between two head robots already in $\overline{\mathcal{R}}$ cannot be resolved and may result in deadlock. Therefore the conflict should be resolved with priority determination in the first place to ensure that there are no deadlocks among head robots.


In the algorithm, even $A_{i}$ is in $\underline{\mathcal{R}}$, if $A_{i}$ is to become a head robot, it has to make sure its masked velocity solved as a head robot is not in conflict with any head robots in $\overline{\mathcal{R}}$, Thus the term $\vect{v}_{A_{i}}^{{\normalfont \overline{\text{m}}}}$ in Fig. \ref{fig:priority-alg} applies to all robots in $\mathcal{R}$ for priority determination. MVO cone is used to detect conflict between the two robots $A_{i} \in \mathcal{R}$ and $A_{j} \in \overline{\mathcal{R}}$ for any time instance from now such that the two robots are guaranteed to be definite deadlock-free in masked velocity space if the masked velocity $\vect{v}_{A_{i}}^{\overline{\text{m}}}$ of $A_i$ is not in $ MVO_{A_{i}|A_{j}}^{\infty}$. Further, even $\vect{v}_{A_{i}}^{\overline{\text{m}}}$ is in $ MVO_{A_{i}|A_{j}}^{\infty}$, if the projection of masked velocity $\vect{v}_{A_{i}}^{\overline{\text{m}}}$ onto masked velocity $\vect{v}_{A_{j}}^{\overline{\text{m}}}$ is in the same direction with $\vect{v}_{A_{j}}^{\overline{\text{m}}}$, or $\vect{v}_{A_{i}}^{{\normalfont \overline{\text{m}}}}\cdot \vect{v}_{A_{j}}^{{\normalfont \overline{\text{m}}}}<0$ equivalently, the deadlock induced by $A_{i}$ and $A_{j}$ is unlikely because the robot behind can follow the one in front. On condition that $\vect{v}_{A_{i}}^{\overline{\text{m}}}$ is in $ MVO_{A_{i}|A_{j}}^{\infty}$ and $\vect{v}_{A_{i}}^{{\normalfont \overline{\text{m}}}}\cdot \vect{v}_{A_{j}}^{{\normalfont \overline{\text{m}}}}\ge0$, if robot $A_{i}$ has a larger accumulative importance $S_{A_{i}}$ than any such $A_{j}$, the algorithm ensures that the deadlock potential is eliminated by $A_{i}$ replacing any such $A_{j}$ into $\overline{\mathcal{R}}$. The introduction of cumulative importance $S_{A_{i}}$ prevents frequent switchings between high and normal priority for each robot and maintains the consistency of deadlock resolving.

%

\subsection{Solving Process}

\begin{figure}[tb]
 \removelatexerror
  \begin{algorithm}[H]
  \setstretch{0.8}
  \small
   \caption{MCCA}
    \hspace*{\algorithmicindent} \textbf{Input}
    $\mathcal{R}$: All robots $\mathcal{O}$: All obstacles\\
    \hspace*{\algorithmicindent} \textbf{Output}
    Robot control inputs \\
    \textit{Initialization}\;
   \textit{Loop Process}\\
   \For(\emph{Control Cycle}){$ each~ A_i \in \mathcal{R}$}
   {
      Sense central position/velocity of $A_{i}$\;
      Sense obstacle line segments\;
      Receive from each $A_{j\neq i}\in\mathcal{R}$ the states  (central position/velocity, radius, masked velocity and priority)\;
      Execute Algorithm 1 for $A_{i}$\;

        Construct $ORCA^{\tau}_{A_i|\mathcal{O}}$\;
        \For{$A_{j\neq i}\in\mathcal{R}$} {
             Construct $ORCA^{\tau}_{A_i|A_j}$\;
             Construct $MCCA^{\tau}_{A_i|A_j}$ if $A_i \in \underline{\mathcal{R}}$\;
        }
        Get $ORCA^{\tau}_{A_i}$ from all $ORCA^{\tau}_{A_i|A_j}$ and $ORCA^{\tau}_{A_i|\mathcal{O}}$\;
        Get $MCCA^{\tau}_{A_i}$ from all $MCCA^{\tau}_{A_i|A_j}$ and $ORCA^{\tau}_{A_i|\mathcal{O}}$ if $A_i \in \underline{\mathcal{R}}$\;
        Get preferred velocity $\vect{v}_{A_i}^{\text{pref}}$\;
        Solve masked velocity $\vect{v}_{A_i}^{\text{m}}$ with holonomic QP\;
        Solve control inputs $\vect{r}_{A_i}$ with nonholonomic QP\;
        Apply control inputs to actuators of $A_i$
   }
\end{algorithm}
\caption{Algorithm for Masked Cooperative Collision Avoidance.}
\label{fig:mcca-alg}
\end{figure}
\setlength{\belowcaptionskip}{-8pt}
By combining the QP formulation and the concept of masked velocity, we summarize the solving process of masked cooperative collision avoidance in Fig. \ref{fig:mcca-alg}. The decision-making process of MCCA is fully decentralized where the state of each robot is available to all.

\section{An example with differential-drive robots}\label{example}
\begin{figure}[tb]
\centerline{\includegraphics[scale=0.17] {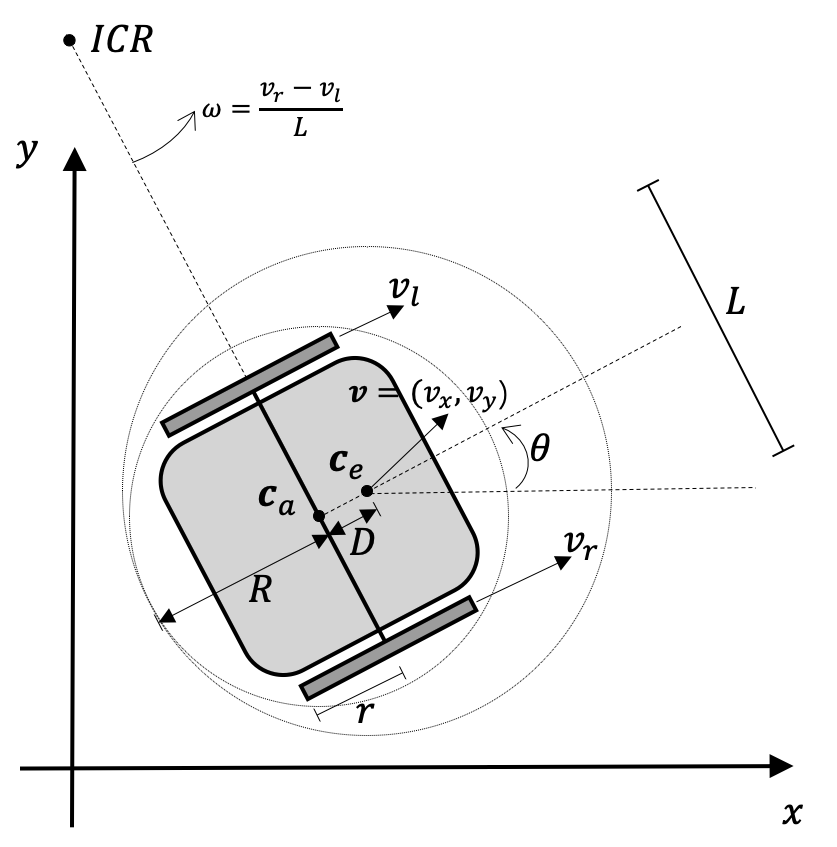}}
\caption{Differential-drive kinematics. Left wheel velocity and right wheel velocity are controlled with separate motors. $r$ is the wheel radius. $L$ is the distance between the wheels. Velocity of the wheel axis center $\vect{c}_a$ is always perpendicular to the axis thus not fully controllable. The effective center $\vect{c}_e$ converts the robot into a disc with radius $R+D$ and $\vect{v} = \left( v_x, v_y \right)$. Instantaneous Center of Rotation (ICR) is introduced for the convenience of calculating angular velocity $\omega$\cite{klancar2017}. We define robot heading as the direction of $\vect{c}_e - \vect{c}_a$ as in the figure.}
\label{fig-dd-kinematics}
\end{figure}

Given the features introduced above, we present in this section a detailed implementation for differential-drive robots. Fig. \ref{fig-dd-kinematics} provides the basic kinematic model for differential-drive robots. The wheel axis center is not fully controllable for this type of robots and we adopt the effective center and radius as in \cite{Kluge2005}. However, it is only necessary to ensure $D>0$ for the sole purpose of kinematic adaption. Instead of choosing the value delicately and resulting in a significantly larger disc radius, we assume $D/R\approx 0.1$ or less, which is much smaller than the recommended value ($D=R$) in \cite{snape2010smooth}. \\

\begin{table}[hbt]
\small
\caption{Variables and parameters for the example}
\begin{center}
\begin{tabular}{|c|c|}
\hline
\textbf{Variable} & \textbf{Description}                \\
\hline
 $v_{\text{l}}$ & Left wheel linear velocity             \\
\hline
$v_{\text{r}}$  & Right wheel linear velocity            \\
\hline
$\vect{v} = \left( v_x, v_y \right)$  & Translational velocity vector of $\vect{c}_e$  \\
\hline
$\delta_{\zeta}$  & Auxiliary variable for half-plane $\zeta$    \\
\hline
$\omega$  & Angular velocity of the robot \\
\hline
$\delta^{\omega}$  & Auxiliary variable for angular velocity\\
\hline
\textbf{Parameter} & \textbf{Description} \\
\hline
 $v^{\text{c}}_{\text{l}}$ & Current left wheel linear velocity\\
\hline
$v^{\text{c}}_{\text{r}}$ & Current right wheel linear velocity\\
\hline
$\theta$  & Current heading orientation   \\
\hline
  $\vect{v}^{\text{pref}}$ & Preferred velocity vector of the robot\\ 
\hline
  $v^{\text{max}}$  & Maximum linear velocity of wheels  \\
\hline
  $a^{\text{max}}$  & Maximum linear acceleration of wheels \\
\hline
   $\varDelta t$ & Time step    \\
\hline
  $\vect{v}_{H}^{\text{opt}} $  & Holonomic optimization velocity   \\
\hline
  $\theta_{H}^{\text{opt}} $ & Holonomic optimization heading   \\
\hline
  $\mu $  & Angular control level    \\
\hline
\end{tabular}
\end{center}
\label{tab-dd-var}
\end{table}

\subsection{The Nonholonomic QP Formulation for Real Velocity}
Equation \eqref{DDform:main} presents the differential-drive QP formulation for solving the real velocity and Table \ref{tab-dd-var} summarizes the variables and parameters it mentions. Its objective function has been adjusted according to Equation \eqref{eq:qp-mcca}. In addition to the basic differential-drive kinematic constraints, a special type of angular control constraint \eqref{DDform:angular} is added as a customized constraint \eqref{Gform:kin2} to solve a problem unique to differential-drive robots, which is introduced in the next subsection. 

Constraint \eqref{DDform:ca} includes the MCCA half-planes for all robots ($\mathcal{M}=\emptyset$ for head robots) and the ORCA half-planes for all robots and obstacles. Constraints \eqref{DDform:kin1} to \eqref{DDform:kin6} describe the kinematic constraints for differential-drive robots. $\vect{v} = \left( v_x, v_y \right)$ is the velocity vector for the effective center, which is away from the axis center with distance $D>0$. Constraints \eqref{DDform:kin1}, \eqref{DDform:kin2} and \eqref{DDform:kin3} establish the direct kinematics where $v_{\text{l}}$ and $v_{\text{r}}$ are the control inputs. Constraints \eqref{DDform:kin4} and \eqref{DDform:kin6} set the maximum velocity and acceleration for the control inputs. For each robot, $a^{\text{max}} \varDelta t$ is the largest possible change of wheel linear velocity between two consecutive control cycles with time step $\varDelta t$.

\begin{subequations}\label{DDform:main}
\begin{align}
& \text{min}  && \begin{aligned}[t] &\alpha _{1} \left \| \vect{v}-\vect{v}^{\text{pref}} \right \| _{2}^{2} + \\
&\alpha _{2}\sum_{i\in \mathcal{O}}^{} \delta_{i}^{2}  + \alpha _{3}\sum_{j\in \mathcal{R}}^{} \delta_{j}^{2} +\alpha _{4}\sum_{k\in \mathcal{M}}^{} \delta_{k}^{2}+\alpha_{5}\delta_{\omega }^{2} \end{aligned}  \tag{\ref{DDform:main}} \\
& \text{s.t.} && \det\begin{pmatrix}
  v_{x} -  p_{x}^{\zeta}&d_{x}^{\zeta} \\[1pt]
   v_{y} - p_{y}^{\zeta} &d_{y}^{\zeta}
\end{pmatrix} \leq \delta_{\zeta}, \zeta\in \mathcal{O}\cup \mathcal{R}\cup \mathcal{M} ,\delta_{\zeta}\geq 0         \label{DDform:ca}  \\
&                && v_{x}=\left ( \frac{\cos\theta }{2} +\frac{D\sin\theta }{L}\right )v_{l}+\left ( \frac{\cos\theta }{2} -\frac{D\sin\theta }{L}\right )v_{r}          \label{DDform:kin1} \\
&                && v_{y}=\left ( \frac{\sin\theta }{2} -\frac{D\cos\theta }{L}\right )v_{l}+\left ( \frac{\sin\theta }{2} +\frac{D\cos\theta }{L}\right )v_{r}          \label{DDform:kin2} \\
&                && \omega =  \frac{v_{r}-v_{l}}{L}         \label{DDform:kin3} \\
&                &&\left | v_{l}  \right | , \left | v_{r}  \right |\leq v^{\text{max}}          \label{DDform:kin4} \\
&                &&\left |v_{l}-v_{l}^{\text{c}}\right |,\left |v_{r}-v_{r}^{\text{c}} \right |\leq a^{\text{max}} \varDelta t         \label{DDform:kin6} \\
&                && \begin{aligned}[t] & \int_{0}^{T}  \left |\omega  \right | -\frac{2 a^{\text{max}}\cdot t}{L} ~dt  \leq \theta'+\delta^{\omega} , \delta^{\omega}\geq 0, \\
& T=\frac{\theta' L}{2 a^{\text{max}}} , \theta' =\left\{\begin{matrix} ~~ \frac{\angle (\theta,\theta_{H}^{\text{opt}})}{\mu} ~~\text{if} ~~ \vect{v}_{\omega}\cdot \vect{v}_{H}^{\text{opt}}>0 \\ \angle (\theta,\theta_H^{\text{opt}}) ~~~~\text{otherwise}\end{matrix}\right.\end{aligned} \label{DDform:angular}
\end{align}
\end{subequations}

\subsection{Angular Control for Heading Oscillation Avoidance}

\begin{figure}[bt]
\centerline{\includegraphics[scale=0.10] {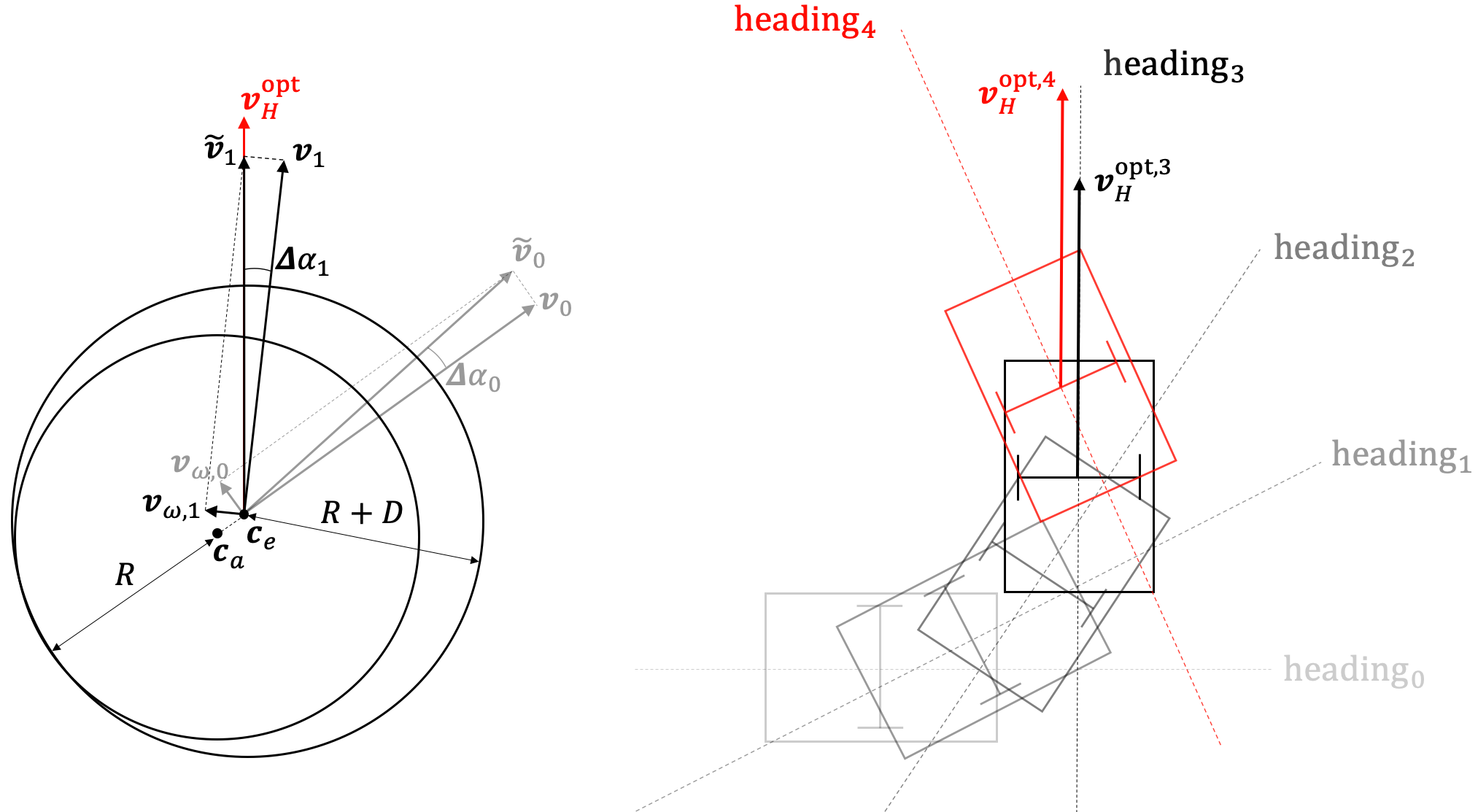}}
\caption{Heading oscillation with differential-drive robots. $\vect{v}_{H}^{\text{opt}}$ is the solution to QP \eqref{DDform:main} considering only constraint \eqref{DDform:ca} at each time step. A nonholonomic robot is not able to reach $\vect{v}_{H}^{\text{opt}}$ instantly but will try to reach it under kinematic constraints. The solved velocity of the robot effective center at each time step can be decomposed into $\widetilde{\vect{v}}= \vect{v}_{\omega}+\vect{v}$ where $\vect{v}_{\omega}$ is perpendicular to robot heading direction and $|\vect{v}_{\omega}| = \omega D$. The overshooting phenomenon is that when $\widetilde{\vect{v}}$ is very close to $\vect{v}_{H}^{\text{opt}}$, the robot may still have a large angular velocity $\omega$ which will divert the heading of robot further away ($\text{heading}_4$) from the target holonomic velocity and its heading ($\text{heading}_3$).} 
\label{fig-dd-overshooting}
\end{figure}

We discovered a heading oscillation issue when $D$ is relatively small and acceleration is taken into consideration. The phenomenon is demonstrated in Fig. \ref{fig-dd-overshooting}. The cause of heading oscillation is twofold. First, we consider a differential-drive robot in an empty environment thus $\vect{v}_{H}^{\text{opt}}=\vect{v}^{\text{pref}}$ and remains unchanged. The angle between $\widetilde{\vect{v}}$ and $\vect{v}$ in Fig. \ref{fig-dd-overshooting} is $\varDelta\alpha=\arctan\frac{\omega D}{|v|}$. A small $D$ will lead to a angle so small that when the angle between $\widetilde{\vect{v}}$ and $\vect{v}_{H}^{\text{opt}}$ is closing to zero, the robot may still have a large angular velocity and overshoot before the control inputs can decelerate $\omega$ to zero, which results in heading oscillation. Second, the unchanged $\vect{v}_{H}^{\text{opt}}$ assumption is not appropriate. It changes constantly with $\vect{v}^{\text{pref}}$ and states of other robots varying. If the newer $\vect{v}_{H}^{\text{opt}}$ is solved to be very close to the current robot heading, the robot will overshoot if its angular velocity cannot be decreased to zero in time. 

\begin{figure}[bt]
\centerline{\includegraphics[scale=0.085] {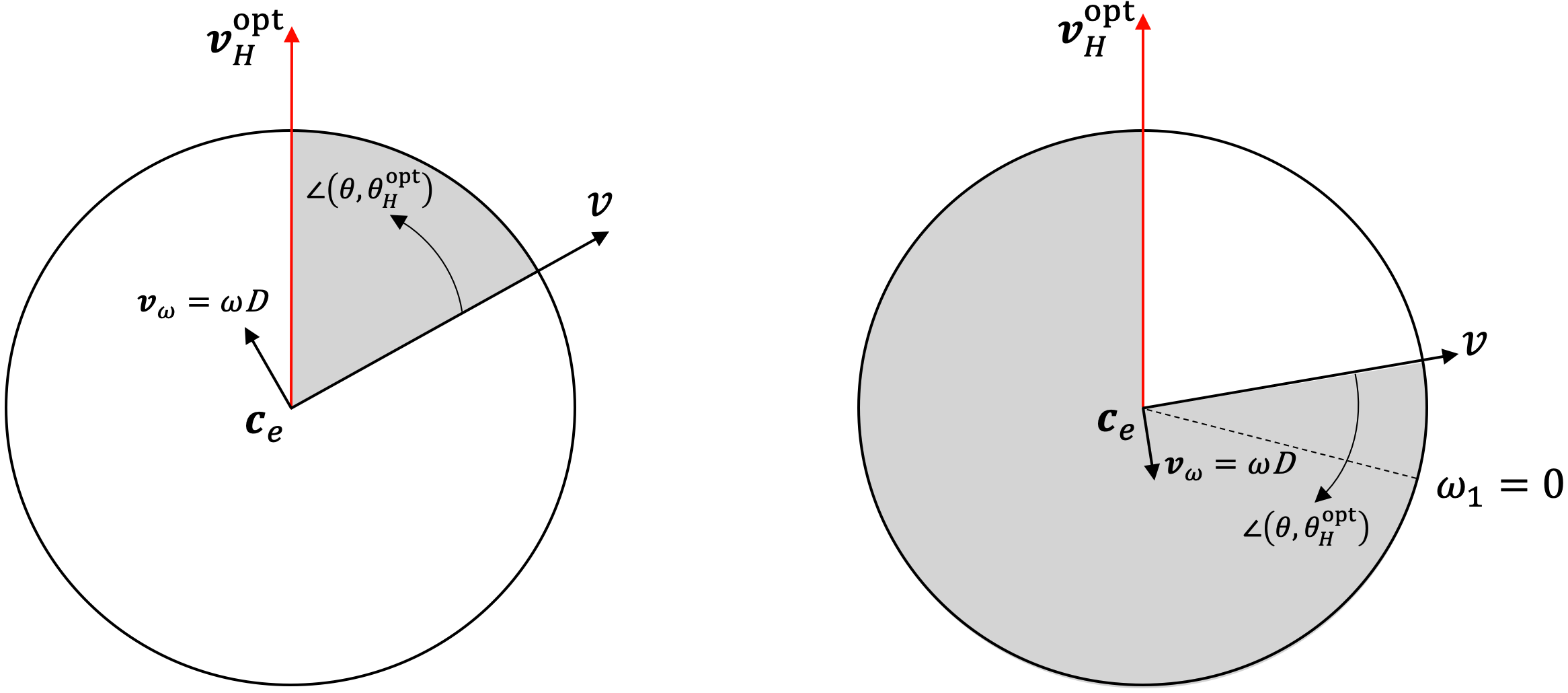}}
\caption{Two cases for $\theta^{'}$. With $\vect{v}_{\omega}\cdot \vect{v}_{H}^{\text{opt}}\le0$ on the right, instead of traveling the shaded angular space as the robot on the left does, the robot may also choose to decrease $\omega$ to zero at the dashed line, and reversely increase $\omega$ to reach $\vect{v}_{H}^{\text{opt}}$ faster (like an S-turn). After $\omega$ reverses, it becomes the first case on the left.}
\label{fig-dd-twocondition}
\end{figure}

To avoid heading oscillation we introduce constraint \eqref{DDform:angular}. Define $\angle (\theta,\theta_{H}^{\text{opt}})$ as the angle that the robot heading $\theta$ needs to turn in the direction of current $\vect{v}_{\omega}$ to reach $\vect{v}_{H}^{\text{opt}}$. Basically, the angular control constraint limits the increase of $\omega$ such that the current $\omega$ is able to decrease to zero using the largest angular acceleration $\frac{2 a^{\text{max}}}{L}$ before the travelled angle reaches the current $\angle (\theta,\vect{v}_{H}^{\text{opt}})$. 

To further reduce the occurrence of the second cause of heading oscillation due to the changing $\vect{v}_{H}^{\text{opt}}$, we divide $\theta^{'}$ by $\mu>1$ when the angle is less than $\pi$ or $\vect{v}_{\omega}\cdot \vect{v}_{H}^{\text{opt}}>0$ equivalently, which will prevent $\omega$ from increasing too much. However, $\theta^{'}$ is not divided by $\mu>1$ in the following situation. When the angle $\angle (\theta,\theta_{H}^{\text{opt}})$ is at least $\pi$, or $\vect{v}_{\omega}\cdot \vect{v}_{H}^{\text{opt}}\le0$ equivalently, the differential-drive robot may choose to perform an S-turn as shown in Fig. \ref{fig-dd-twocondition}. When $\angle (\theta,\theta_{H}^{\text{opt}})$ is closer to $2\pi$ the robot is more likely to perform such an S-turn maneuver. Thus we don't reduce $\theta^{'}$ by $\mu>1$ in this case.

\section{Experimentation}\label{exp}
We describe the implementation and simulation results of MCCA with differential-drive robots in various scenarios.

\subsection{Implementation Details}


All robots are differential-drive with \SI[mode=text]{950}{mm} $\times$ \SI[mode=text]{725}{mm} $\times$ \SI[mode=text]{285}{mm} in three dimensions. Each robot is equipped with two powered wheels of radius $r=\SI[mode=text]{0.1}{m}$ and two passive caster wheels to balance the robot. The maximum linear velocity is \SI[mode=text]{2}{m/s} and the maximum linear acceleration is \SI[mode=text]{2}{m/s^2}. Each powered wheel is controlled by a servomotor, which given the control inputs will power the wheel with up to the maximum linear acceleration. The distance between $\vect{c}_a$ and $\vect{c}_e$ is $D= \SI[mode=text]{0.015}{m}$. Robot bounding circle radius $R=\SI[mode=text]{0.485}{m}$. The effective radius is $R+D = \SI[mode=text]{0.5}{m}$, which is only 3 percent larger than $R$. The time step for each control cycle is $\varDelta t = \SI[mode=text]{0.25}{s}$. The time horizon is $\tau= \SI[mode=text]{17}{s}$. The weight parameters in Equation \eqref{DDform:main} are $\alpha_1=10^{-2}$, $\alpha_2=10^{4}$, $\alpha_3=10^{2}$, $\alpha_4= 1$, $\alpha_5=2\times 10^{4}$. The angular control parameter $\mu=9$. The initial tabu steps $\eta = 30$.

In each simulation scenario, a broadcast service is used to receive the states of all the robots. The states are broadcast to all the robots in the system. Each robot at each time step will use its own state and the broadcast states of other robots to determine its control inputs. For the purpose of this research, the simulation does not include the sensing and localization part, which is typically accomplished in the real world with Simultaneous Localization and Mapping (SLAM). However, we enforce a uniform-distributed noise of location and angle for each robot, with an amplitude of 0.01 m and 1 degree, respectively. We apply a simple method to set the preferred velocity $\vect{v}^{\text{pref}}$ of a robot to point to the goal from its current location with magnitude no more than the maximum velocity. Webots \cite{michel2004}, a mobile robotics simulation software, is used for our simulation. We use PROX-QP \cite{bambade2022} for solving the proposed QP problem in the simulation. 

We design the simulation scenarios such that they should be able to verify the capability of MCCA in two aspects. First, collision avoidance is the minimum requirement. Second, deadlock avoidance is stable and efficient, and is functional even with scenarios where deadlocks are easy to occur. Besides, we also verify the effectiveness of angular control for differential-drive robots to avoid heading oscillation.

\subsection{Simulation Results} 

\setlength{\belowcaptionskip}{0pt}
\begin{figure}[htbp]
     \begin{subfigure}{0.5\textwidth}
         \centering
         \includegraphics[width=0.95\textwidth]{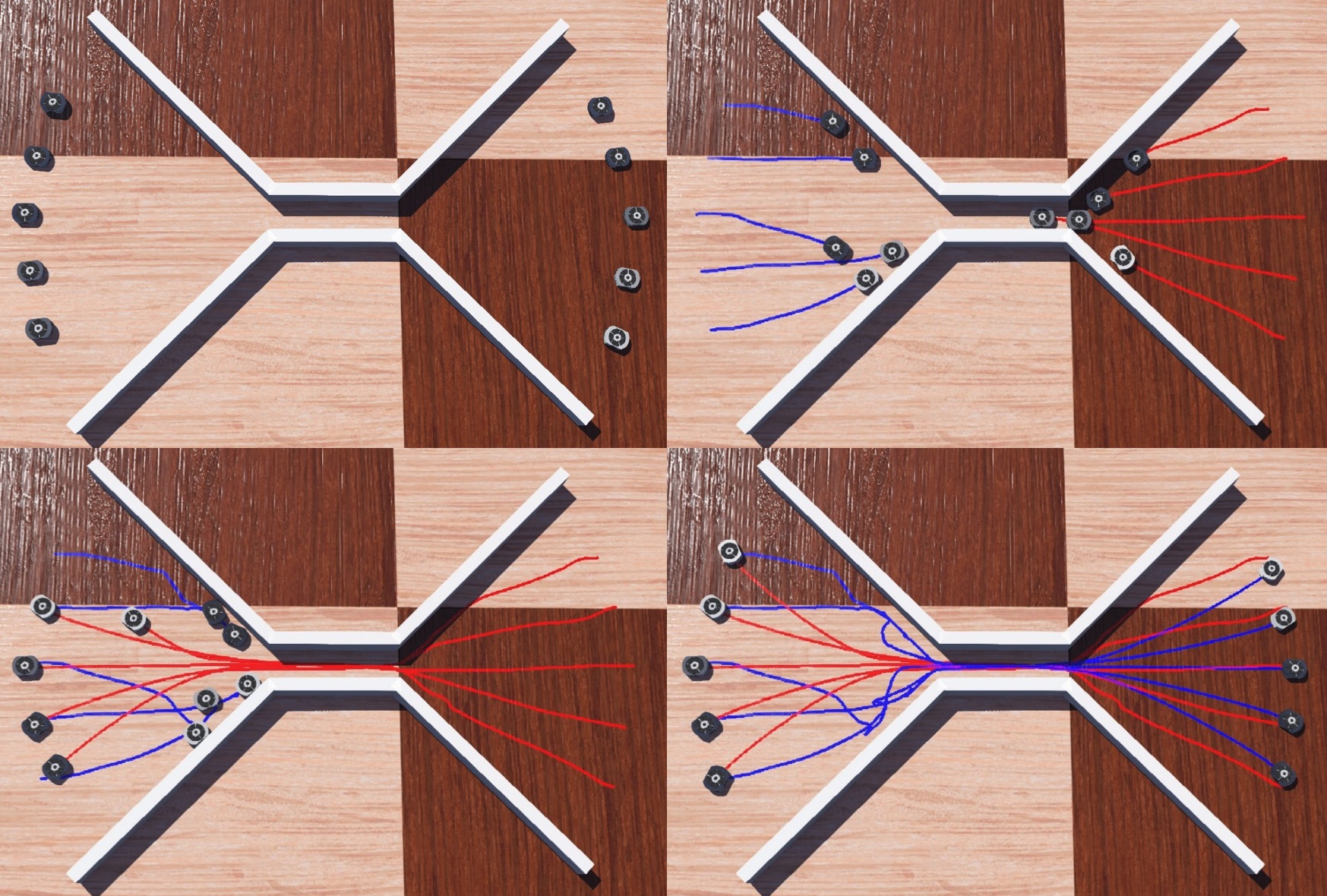}
         \caption{}
         \label{fig:scen1}
     \end{subfigure}
     \begin{subfigure}{0.5\textwidth}
        \centering
         \includegraphics[width=0.95\textwidth]{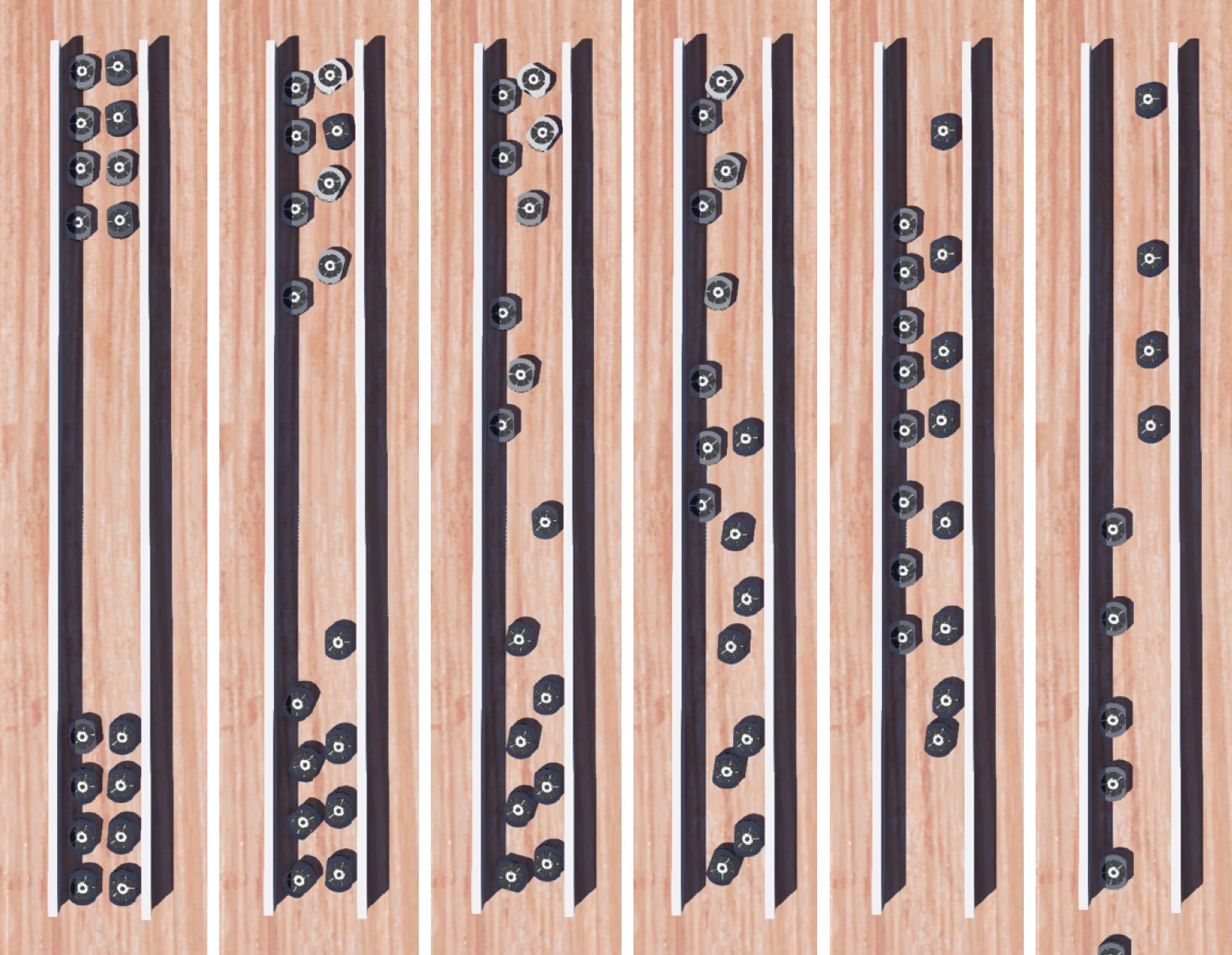}
         \caption{}
         \label{fig:scen5}
     \end{subfigure}
          \begin{subfigure}{0.5\textwidth}
       	\centering
         \includegraphics[width=0.95\textwidth]{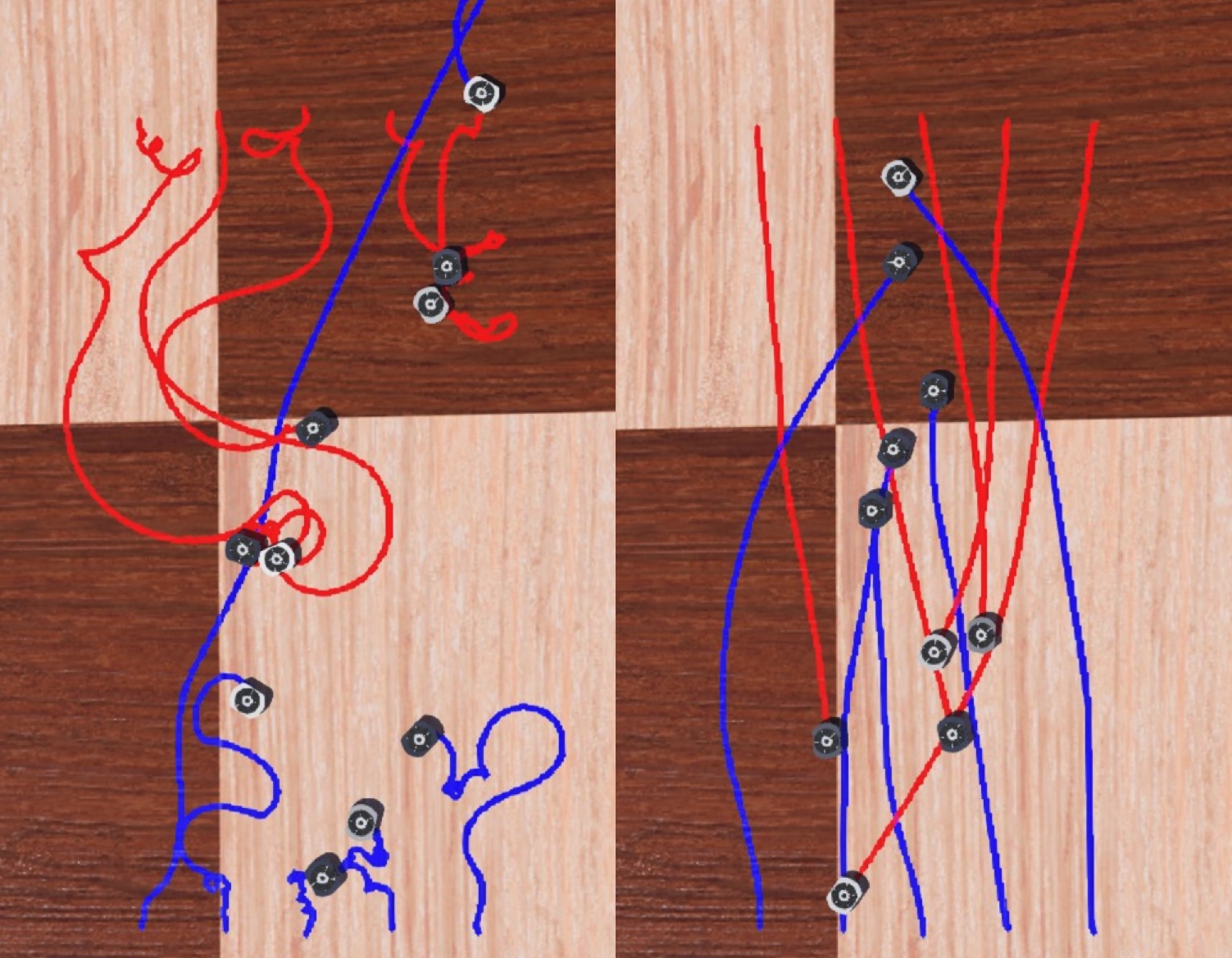}
         \caption{}
         \label{fig:scen7}
     \end{subfigure}
     \caption{Selected simulation results. (a) Scenario 1: The 5 robots on one side overtake the positions of the 5 robots on the other side with a one-lane passage. (b) Scenario 2: The 8 robots on each sides of a two-lane passage try to reach the other side. (c) Scenario 7: Before and after introducing angular control for differential-drive robots with poor maneuverability, whose maximum linear acceleration is reduced to \SI[mode=text]{0.2}{m/s^2} from \SI[mode=text]{2}{m/s^2}.}     
     \label{fig:scen}
\end{figure}

We introduce the scenarios briefly. 
\begin{itemize}
\item Scenario 1: Two groups each with five robots overtaking the positions on the other side with a one-lane passage. 
\item Scenario 2: Two groups each with eight robots traveling in a two-lane passage and give way spontaneously and smoothly. 
\item Scenario 3: Ten robots taking repeated round-trips with multiple one-lane passages. 
\item Scenario 4: Twenty robots taking repeated round-trips in an environment with densely placed obstacles.
\item Scenario 5: Forty robots taking repeated round-trips in a confined and congested area.
\item Scenario 6: Thirty-four robots converging into a passage. 
\item Scenario 7: Ten robots with poor maneuverability taking repeated round-trips. 
\end{itemize}

All the listed scenarios are demonstrated in the attached video. Scenario 3, 4, 5, and 7 involve repeated round-trips and remain collision-free and deadlock-free for hundreds of simulation hours and continue to be so. So does Scenario 1 if robots continuously overtake positions on the other side. Selected results from scenarios 1, 2 and 7 are summarized as follows. 

For Scenario 1, Fig. \ref{fig:scen1} shows the trace of each robot. If robots from both sides rush into the passage, deadlock could happen and some robots will move back and forth for a long period of time. Our method is able to avoid the deadlock situation locally and spontaneously. Some robots will detour or wait outside the passage to give way to the robot in the passage. For Scenario 2, Fig. \ref{fig:scen5} provides a series of screenshots when 8 robots on each side encounter in a two-lane passage. Not only do the robots give way to robots on the other side with what appears to be a zipper merge, but they also complete the process smoothly without central control or sudden change of path. For Scenario 7, Fig. \ref{fig:scen7} shows that angular control is effective to handle robots with low acceleration compared to the maximum velocity. Without angular control, we can observe sudden stops, heading oscillation, and circling issues from these robots. The problems are tackled effectively by angular control. Moreover, the radius of robots are only enlarged marginally, allowing these robots with poor maneuverability to travel smoothly even in congested area and narrow passages. 

\section{Conclusion}\label{conclude}
In this paper, we first provide a multi-objective general QP formulation for solving collision avoidance in a multi-robot system. The formulation is a well-defined convex problem that is easy to solve, applicable to most controllable wheeled mobile robots and linear customized constraints can be incorporated. On top of that, we propose the MCCA method to effectively avoid and resolve deadlocks while minimizing the global coordination to maintain a fully decentralized decision-making mechanism. We define masked velocity and MCCA half-planes mathematically to carry out the intention that enables robots to resolve deadlock situations locally and collectively. The general QP works with MCCA to ensure a collision-free, deadlock-free and smooth path for each robot. To illustrate that, we provide a detailed implementation for differential-drive robots and solve the heading oscillation issue by adding a customized angular control constraint into the QP formulation, which allows robots with poor maneuverability to travel smoothly in the simulation without significantly enlarging the effective radius. The simulation shows very promising results for deadlock avoidance in complex and highly difficult environment, which persuades us to believe that the MCCA method has great advantage and potential in real-world applications.

\bibliographystyle{IEEEtran}
\bibliography{IEEEabrv, ../Literature/ORCA}

\end{document}